\documentclass{article}

\usepackage{PRIMEarxiv}

\usepackage[utf8]{inputenc} 
\usepackage[T1]{fontenc}    
\usepackage{hyperref}       
\usepackage{url}            
\usepackage{booktabs}       
\usepackage{amsfonts}       
\usepackage{nicefrac}       
\usepackage{microtype}      
\usepackage{lipsum}
\usepackage{fancyhdr}       
\usepackage{graphicx}       
\graphicspath{{media/}}     
\usepackage{graphicx}
\usepackage{subcaption}
\usepackage{amsmath}
\graphicspath{{media/}}     
\usepackage{appendix}

\title{Class Conditional Time Series Generation with Structured Noise Space GAN}
\author{
  Hamidreza Gholamrezaei \\
  University of Kaiserslautern-Landau \\
  Kaiserslautern,Germany\\
  \texttt{h\_gholamre17@cs.uni-kl.de} \\
   \And
  Alireza Koochali\thanks{Corresponding author}\\
  IAV GmbH \\
  Kaiserslautern,Germany\\
  \texttt{alireza.koochali@iav.de} \\
   \And
  Andreas Dengel \\
  DFKI GmbH \\
  University of Kaiserslautern-Landau \\
  Kaiserslautern,Germany\\
  \texttt{andreas.dengel@dfki.de} \\
   \And
  Sheraz Ahmed \\
  DFKI GmbH \\
  Kaiserslautern,Germany\\
  \texttt{sheraz.ahmed@dfki.de} \\
}

\begin{document}
\maketitle

\begin{abstract}
This paper introduces Structured Noise Space GAN (SNS-GAN), a novel approach in the field of generative modeling specifically tailored for class-conditional generation in both image and time series data. It addresses the challenge of effectively integrating class labels into generative models without requiring structural modifications to the network. The SNS-GAN method embeds class conditions within the generator’s noise space, simplifying the training process and enhancing model versatility. The model's efficacy is demonstrated through qualitative validations in the image domain and superior performance in time series generation compared to baseline models. This research opens new avenues for the application of GANs in various domains, including but not limited to time series and image data generation.
\end{abstract}

\section{Introduction}

Class-conditional generative models represent significant progress in machine learning, empowering tasks that necessitate controlled data synthesis. Contrary to their unconditional counterparts, these models synthesize samples conditioned on specific class labels, facilitating a more focused generative process. This conditional generation enables the production of class-representative data, which has a multitude of practical applications. For example, such models can augment the diversity of a class in a training dataset, thereby enhancing the robustness of classifiers in subsequent tasks.
\newline\newline\noindent
Moreover, Generative Adversarial Networks (GANs)~\cite{goodfellow_generative_2014} stand to gain from integrating class labels during the training phase. The inclusion of such labels steers the generation process towards enhanced quality and diversity. These labels enrich the model's understanding of the data distribution, allowing the generator to create samples that are not merely plausible but also class-specific. This specificity potentially leads to outputs of higher fidelity and greater variety. Concurrently, the discriminator benefits from class labels by providing a more nuanced task of discriminating between genuine and generated data, evaluating not only the authenticity but also the class alignment, thereby furnishing a more robust learning signal for the generator.
\newline\newline\noindent
Nevertheless, incorporating class labels as a condition into the GAN architecture is crucial for the effective functioning of Conditional GANs (CGANs). The field lacks a unified approach for the integration of conditional information into GAN architectures, leaving researchers to adopt a trial-and-error methodology to identify optimal strategies for condition integration. Scientific efforts have been bifurcated into two primary streams based on discriminator conditioning: classifier-based GANs~\cite{odena_conditional_2017,gong_twin_2019,kang_contragan_2020,zhou_omni-gan_2021,hou_conditional_2022,kang_rebooting_2021} and projection-based GANs~\cite{miyato_spectral_2018,brock_large_2019,miyato_cgans_2018,han_dual_2021}. These categorizations reveal the scarcity of options for conditioning generators. The original CGAN proposal~\cite{mirza_conditional_2014} suggests concatenating one-hot encoded class labels with the noise vector. However, this method may lead to the generator disregarding the condition vector when the number of classes is small or negatively impacting generator performance when the number of classes is large. Another prevalent approach, particularly in the image domain, is the injection of class label information into the generator's batch normalization layer~\cite{dumoulin_learned_2017}. Although successful in the image domain. The batch normalization is not utilized for time series data due to the adverse effects of it on the time-dependent correlation between consecutive steps. Thus, this method is not applicable to time series data.
\newline\newline\noindent
To mitigate these challenges, we introduce Structured Noise Space GAN (SNS-GAN), a novel methodology for projecting class-conditional information into the generator without requiring alterations to the network structure. Our approach is agnostic to network architecture, rendering it compatible with both image and time series data.
\newline\newline\noindent
The subsequent section delineates the proposed SNS-GAN model in detail. Following that, we apply the method to the image domain to provide a qualitative validation of our approach. Lastly, we demonstrate the model's superior efficacy in the time series domain compared to baseline models.

\section{Structured Noise Space GAN (SNS-GAN)}

\begin{figure}[ht]
    \centering
    \includegraphics[width=\textwidth]{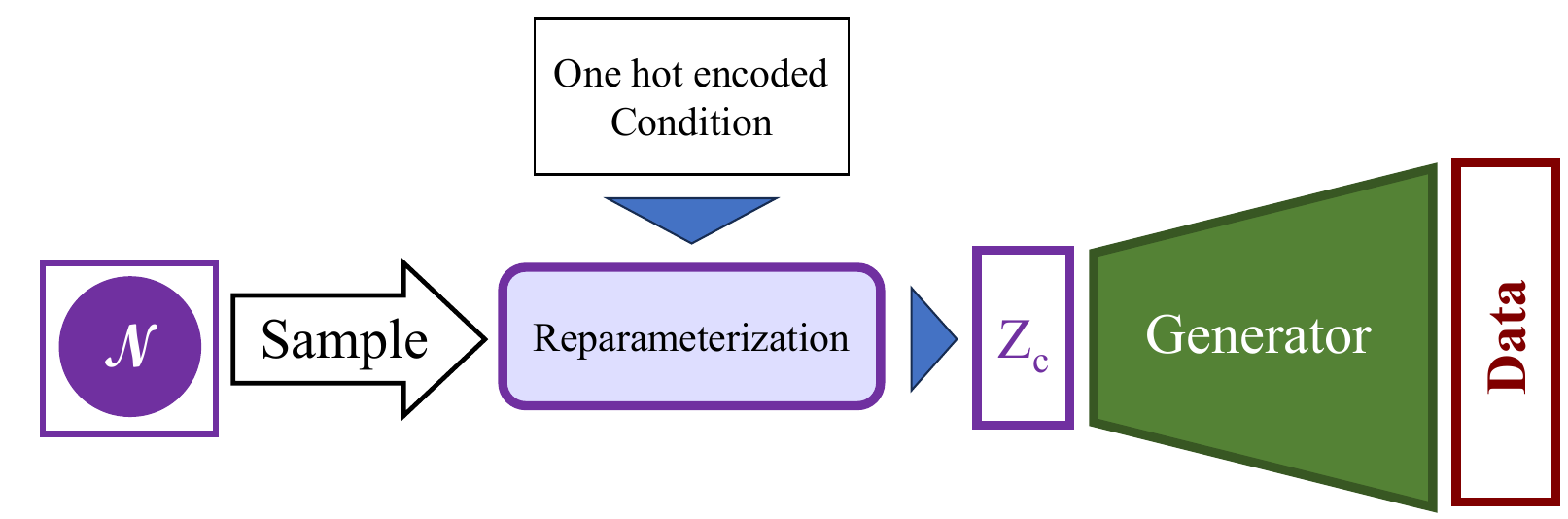}
    \caption{The modified data pipeline for the SNS-GAN generator.}
    \label{fig:sns_gan_gen}
\end{figure}

In the realm of generative modeling, the ability to generate data specific to a given class out of multiple classes in a dataset is a significant milestone. Typically, data belonging to $N$ classes exhibit $N$ modes, with each mode corresponding to a particular class. The challenge lies in synthesizing samples that not only look realistic but also adhere to the specified class label. This requires the generative model to correctly map the input noise to the appropriate data mode, leveraging the class label information.
\newline\newline\noindent
Current methods in Generative Adversarial Networks (GANs) typically involve sampling noise from a standard Gaussian distribution and then conditioning the generator with explicit class labels to produce class-specific outputs. However, this study introduces an innovative approach that deviates from the norm by proposing a structured noise space for GANs. This structured noise space is multimodal, with each mode directly linked to a data class. By sampling noise from this structured distribution, the noise vector inherently carries the class information, eliminating the need for explicit class labels during generation.
\newline\newline\noindent
To actualize the SNS-GAN, we utilize an $N$-dimensional Gaussian distribution to represent the noise space, where each dimension is paired with a data class. When generating a noise vector for a specific class, we adjust the mean of the corresponding dimension to a non-zero value while the other dimensions remain centered around zero. This shift in the mean value implicitly encodes the class information within the noise vector itself, guiding the generator to produce a sample from the intended class.
\newline\newline\noindent
The process of generating structured noise vectors, denoted as $z_c$, is facilitated by the reparameterization technique. Initially, a noise vector $z$ is drawn from an $N$-dimensional standard Gaussian distribution. The mean of the vector is then altered based on the one-hot encoding of the target class $c$:

\begin{equation}
    z_c = z + \text{one\_hot\_encode}(c)
\end{equation}
\noindent
This strategy simplifies the sampling process from the structured noise space and enhances the efficiency of GAN training. Figure~\ref{fig:sns_gan_gen} illustrates the modified data pipeline for the generator within the SNS-GAN framework. By supplying the generator with the structured noise vector $z_c$, it learns to map this input to the corresponding class-specific sample without necessitating alterations to the generator's internal architecture or imposing any constraints on its structure.
\newline\newline\noindent
As a result, the generator inherently understands the class conditions and operates effectively as if it were an unmodified, unconditional GAN. This novel approach not only simplifies the training process but also opens up new possibilities for the application of GANs across various domains where class-specific data generation is essential.
\newline\newline\noindent
The SNS-GAN is trained using the original GAN objective:

\begin{equation}
    \min _{G} \max _{D} V(D, G) = \mathbb{E}_{x \sim P_{\text {data }}(x)}[\log (D(x))] + \mathbb{E}_{z \sim P_{\text {noise }}(z)}[\log (1-D(G(z)))].
\end{equation}

In all experiments carried out in this study, the proposed model is trained using Adam optimizer with the learning rate set to 0.0002 and $\beta_1 = 0.5$ and $\beta_2=0.99$.

\section{Experiment I: Proof of Concept on Image Domain}

The inception of SNS-GAN is rooted in the need for a robust method to generate time series data. However, the complex and non-intuitive nature of time series data poses significant challenges for qualitative analysis of new generative approaches. Although quantitative measures exist for evaluating GANs, their limitations necessitate a qualitative assessment to gauge the model's generative capabilities fully. Consequently, we first apply SNS-GAN to the image domain, where visual inspection can intuitively validate the model's effectiveness. This experiment serves as a foundation for subsequent applications to time series data.

\subsection{Datasets}
\subsubsection{MNIST}
To evaluate SNS-GAN, we utilized the MNIST dataset~\cite{lecun_mnist_2010}, which comprises grayscale images of handwritten digits (0--9). Each image is assigned a label within the range [0--9]. The dataset consists of 60,000 training and 10,000 testing images, providing a standard benchmark for assessing generative models.

\subsubsection{CIFAR10}
For a more challenging scenario, we employed the CIFAR-10 dataset~\cite{krizhevsky_learning_2009}, featuring color images across ten distinct classes, including various natural scenes and objects. With its 50,000 training and 10,000 test images, CIFAR-10 escalates the complexity for generative models and serves as an ideal candidate to test SNS-GAN's capabilities.

\subsection{Results and Discussion}
\begin{figure}[ht]
    \centering
    \includegraphics[width=\textwidth]{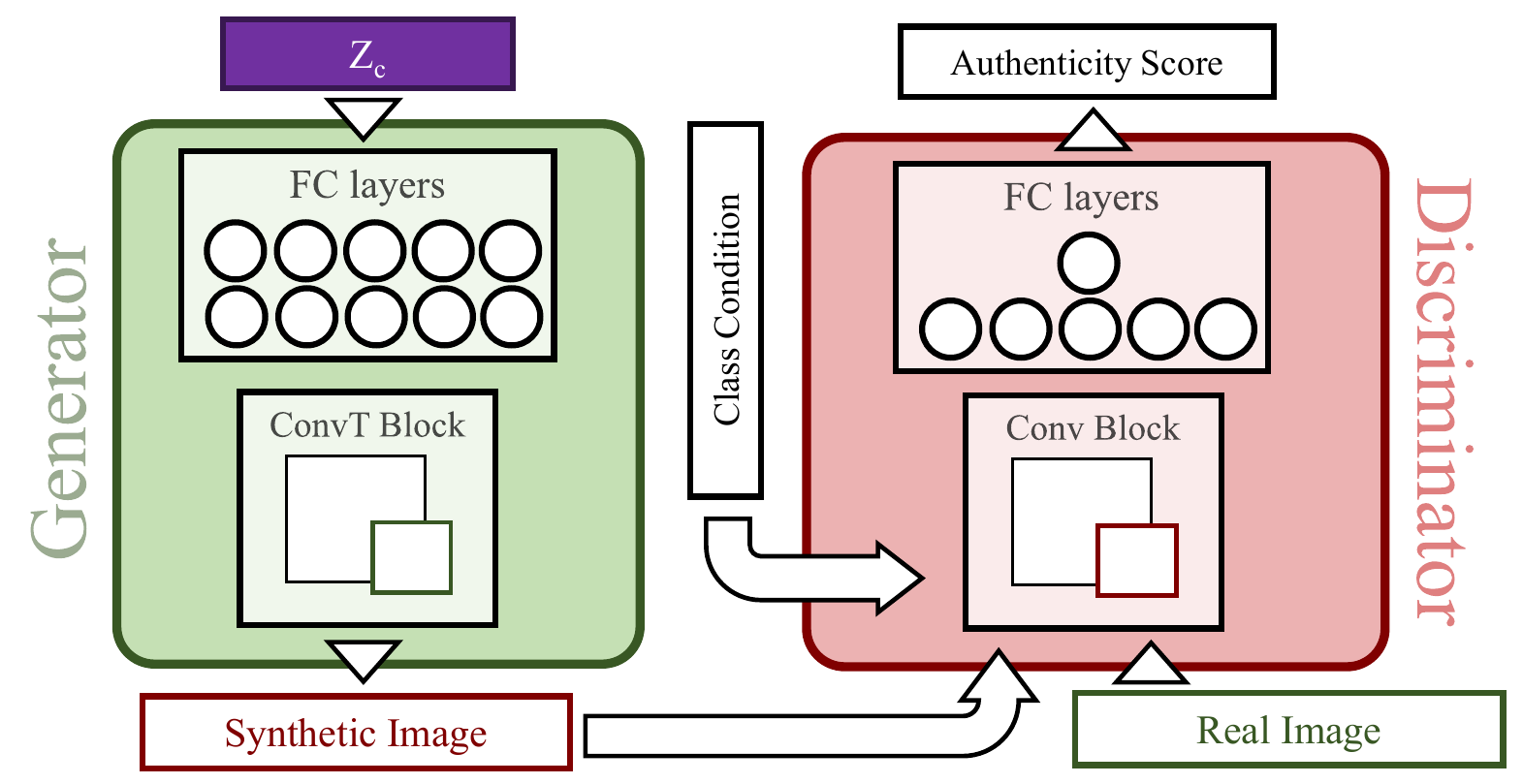}
    \caption{The SNS-GAN architecture for image domain experiments, featuring transposed convolution in the generator and 2D convolution in the discriminator.}
    \label{fig:sns-img}
\end{figure}

\begin{figure}[htbp]
    \centering 
    \begin{subfigure}{0.48\textwidth}
        \includegraphics[width=\textwidth]{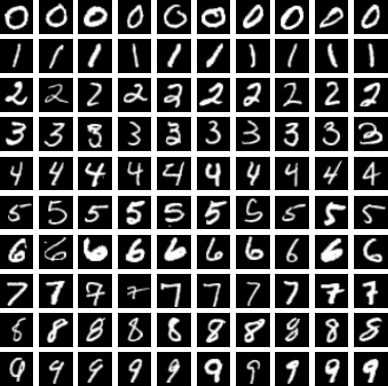}
        \caption{Real MNIST data}
        \label{subfig:real_mnist}
    \end{subfigure}
    \hfill
    \begin{subfigure}{0.48\textwidth}
        \includegraphics[width=\textwidth]{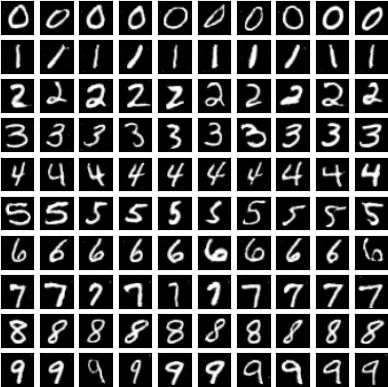}
        \caption{SNS-GAN generated MNIST data}
        \label{subfig:gen_mnist}
    \end{subfigure}
    \caption{Comparison of real and SNS-GAN generated samples from the MNIST dataset.}
    \label{fig:mnist}
\end{figure}

\begin{figure}[htbp]
    \centering 
    \begin{subfigure}{0.48\textwidth}
        \includegraphics[width=\textwidth]{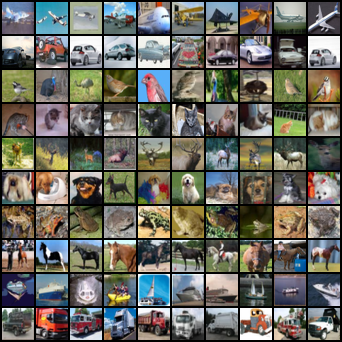}
        \caption{Real CIFAR-10 data}
        \label{subfig:real_cifar10}
    \end{subfigure}
    \hfill
    \begin{subfigure}{0.48\textwidth}
        \includegraphics[width=\textwidth]{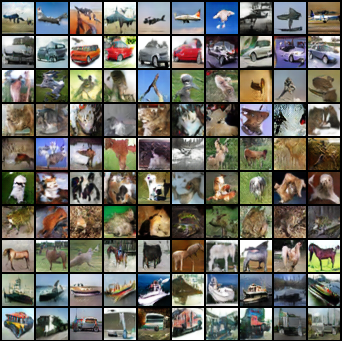}
        \caption{SNS-GAN generated CIFAR-10 data}
        \label{subfig:gen_cifar10}
    \end{subfigure}
    \caption{Comparison of real and SNS-GAN generated samples from the CIFAR-10 dataset.}
    \label{fig:cifar10}
\end{figure}

Figure~\ref{fig:sns-img} illustrates the architectural configuration of SNS-GAN for both the MNIST and CIFAR-10 datasets, highlighting the generator's transposed convolution operations and the discriminator's 2D convolutions. The hyperparameters employed in our experiments are detailed in Appendix~\ref{app:hp}. 
\newline\newline\noindent
Visual comparisons between generated samples and real dataset images are depicted in Figures~\ref{fig:mnist} and \ref{fig:cifar10} for the MNIST and CIFAR-10 datasets, respectively. The results suggest that SNS-GAN successfully enforces class-specific conditions on the generator. The samples display not only high fidelity but also considerable diversity within each class.
\newline\newline\noindent
Figure~\ref{fig:cifar10} presents generated CIFAR-10 samples alongside real ones. Despite the complexity introduced by color and natural scene content, SNS-GAN demonstrates proficient conditional enforcement. Notably, diversity within classes is also preserved. CIFAR-10's low resolution complicates visual fidelity assessment, prompting us to employ Inception Score (IS)~\cite{salimans_improved_2016} and Fréchet Inception Distance (FID)~\cite{heusel_gans_2017} for a quantitative evaluation. These metrics are standard in the image domain for evaluating the quality and diversity of GAN outputs. The comparison with 13 other established GAN models presented in table~\ref{tab:sns_gan_img_res} suggests that SNS-GAN, even with its relatively simple structure, ranks competitively, particularly in terms of FID.
\begin{table}[htbp]
    \centering
    \caption{Quantitative results for CIFAR10 comparing the proposed model with 13 other GAN models.}
    \label{tab:sns_gan_img_res}
    \begin{tabular}{lcc}
    \hline & IS & FID \\
    \hline FCGAN & 6.41 & 42.6 \\
     BEGAN & 5.62 & - \\
     PROGAN & 8.80 & - \\
     LSGAN & 6.76 & 29.5 \\
     DCGAN & 6.69 & 42.5 \\
     WGAN-GP & 8.21 & 21.5 \\
     SN-GAN & 8.43 & 18.8 \\
     Geometric GAN & - & 27.1 \\
     RGAN & - & 15.9 \\
     ACGAN & 8.25 & - \\
     BigGAN & $\mathbf{9.22}$ & $\mathbf{14.7}$ \\
     RealnessGAN & - & 34.6 \\
     SS-GAN & - & 15.7 \\
    \hline SNS-GAN & 6.9 & 14.46 \\
    \hline
    \end{tabular}
\end{table}
\newline\newline\noindent
These findings affirm SNS-GAN's utility for implicit class conditioning and pave the way for its application in the time series domain. The following sections will extend the methodology to time series data, underscoring the potential of SNS-GAN for generating class-conditional sequences.

\section{Experiment II: Time Series Domain}

This section delves into the application of the SNS-GAN model to the generation of class-conditional time series data. We aim to discern whether the SNS-GAN generator can distinguish between classes in the time series domain without explicit class labels. This inquiry involves outlining the datasets and reference models, detailing the architecture of the proposed framework, and presenting the experimental findings.

\subsection{Datasets}

Our evaluation employs four datasets from the UCR benchmark~\cite{chen_ucr_2015}, a repository of 128 univariate time series datasets. The selection was motivated by the diversity in the number of classes and the length of time steps, offering a comprehensive examination of model performance under varied conditions. The chosen datasets, enumerated in Table~\ref{tab:ts_ds_prop}, represent a range of feature combinations to test the models' adaptability.
\begin{table}
    \centering
    \caption{Dataset employed for time series experiments alongside their key characteristics}
    \label{tab:ts_ds_prop}
    \begin{tabular}{llllll}
    \hline \textbf{Feature} & \textbf{Name} & \textbf{Class} & \textbf{Length} & \textbf{Train} & \textbf{Test} \\
    \hline \begin{tabular}{l} 
    low class number \\
    low time step length
    \end{tabular} & Smooth Subspace & 3 & 15 & 150 & 150 \\
    \hline \begin{tabular}{l} 
    low class number \\
    high time step length
    \end{tabular} & Strawberry & 2 & 235 & 613 & 370 \\
    \hline \begin{tabular}{l} 
    high class number \\
    low time step length
    \end{tabular} & Crop & 24 & 46 & 7200 & 16800 \\
    \hline \begin{tabular}{l} 
    high class number \\
    high time step length
    \end{tabular} & Fifty Words & 50 & 270 & 450 & 455 \\
    \hline
    \end{tabular}
\end{table}
\begin{figure}[htbp]
    \centering 
    \begin{subfigure}{0.48\textwidth}
        \includegraphics[width=\textwidth]{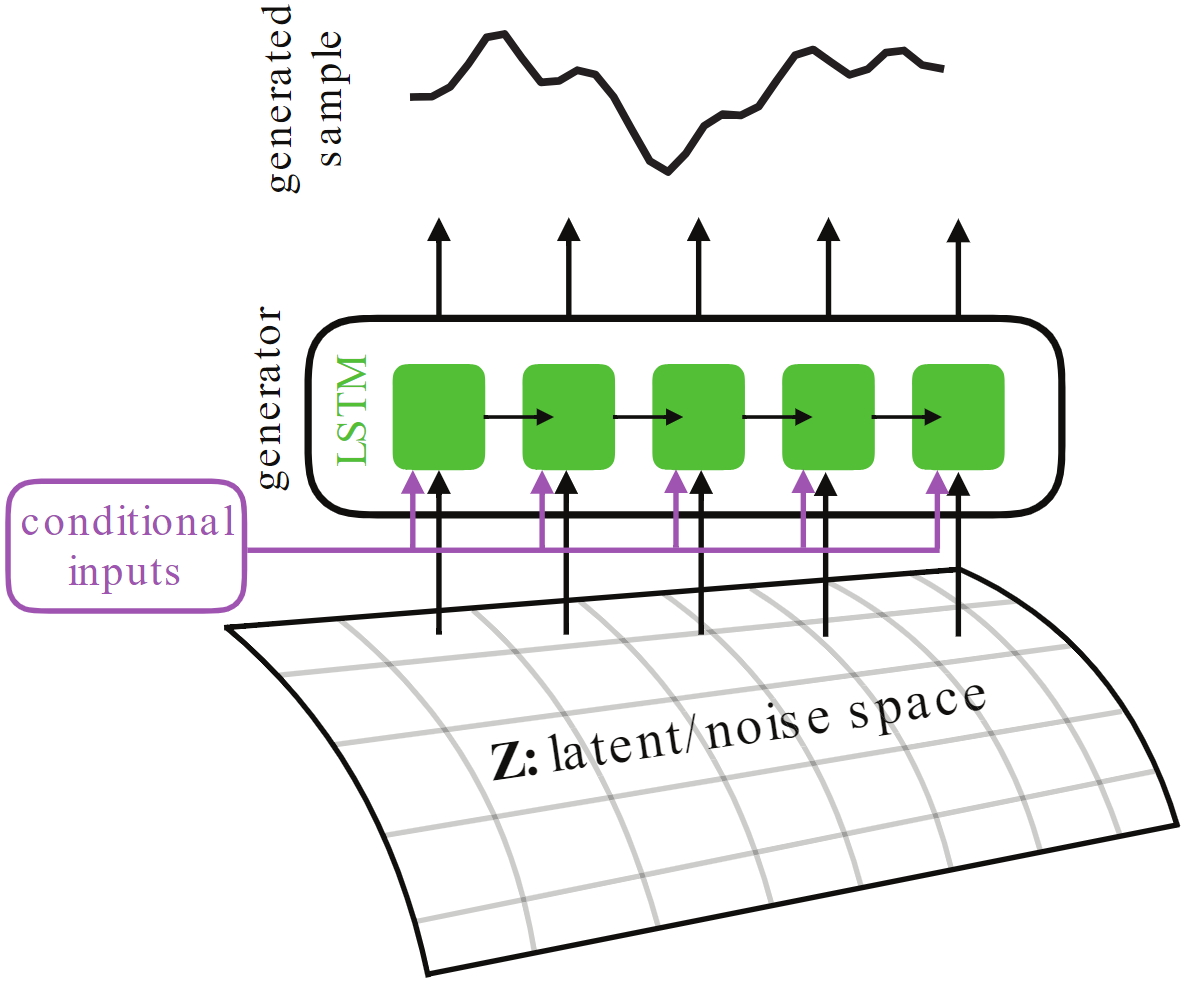}
        \caption{The RCGAN generator}
        \label{fig:rcgan_gen}
    \end{subfigure}
    \hfill
    \begin{subfigure}{0.48\textwidth}
        \includegraphics[width=\textwidth]{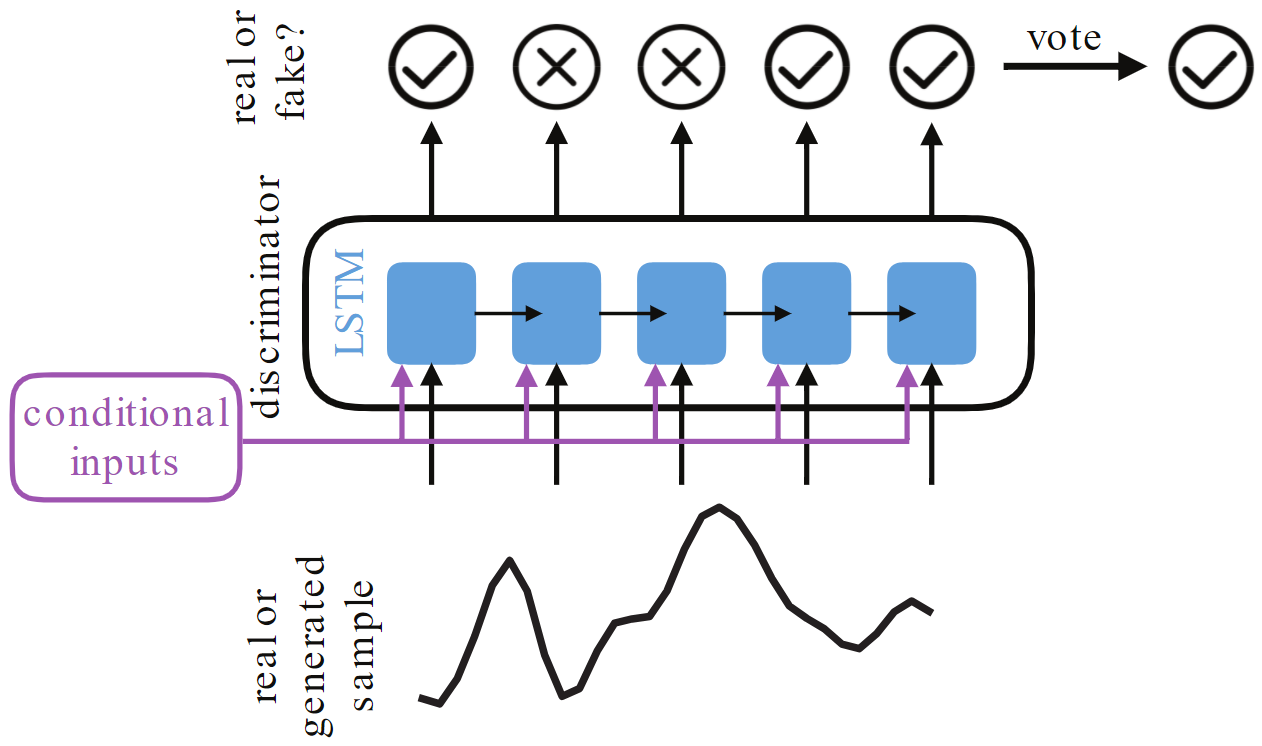}
        \caption{The RCGAN discriminator}
        \label{fig:rcgan_dis}
    \end{subfigure}
    \caption{Schematic of condition imposing pipeline in RCGAN. Figures are adopted from~\cite{esteban_real-valued_2017} }
    \label{fig:rcgan}
\end{figure}
\subsection{Baseline}
\begin{figure}
    \centering
    \includegraphics[width=\textwidth]{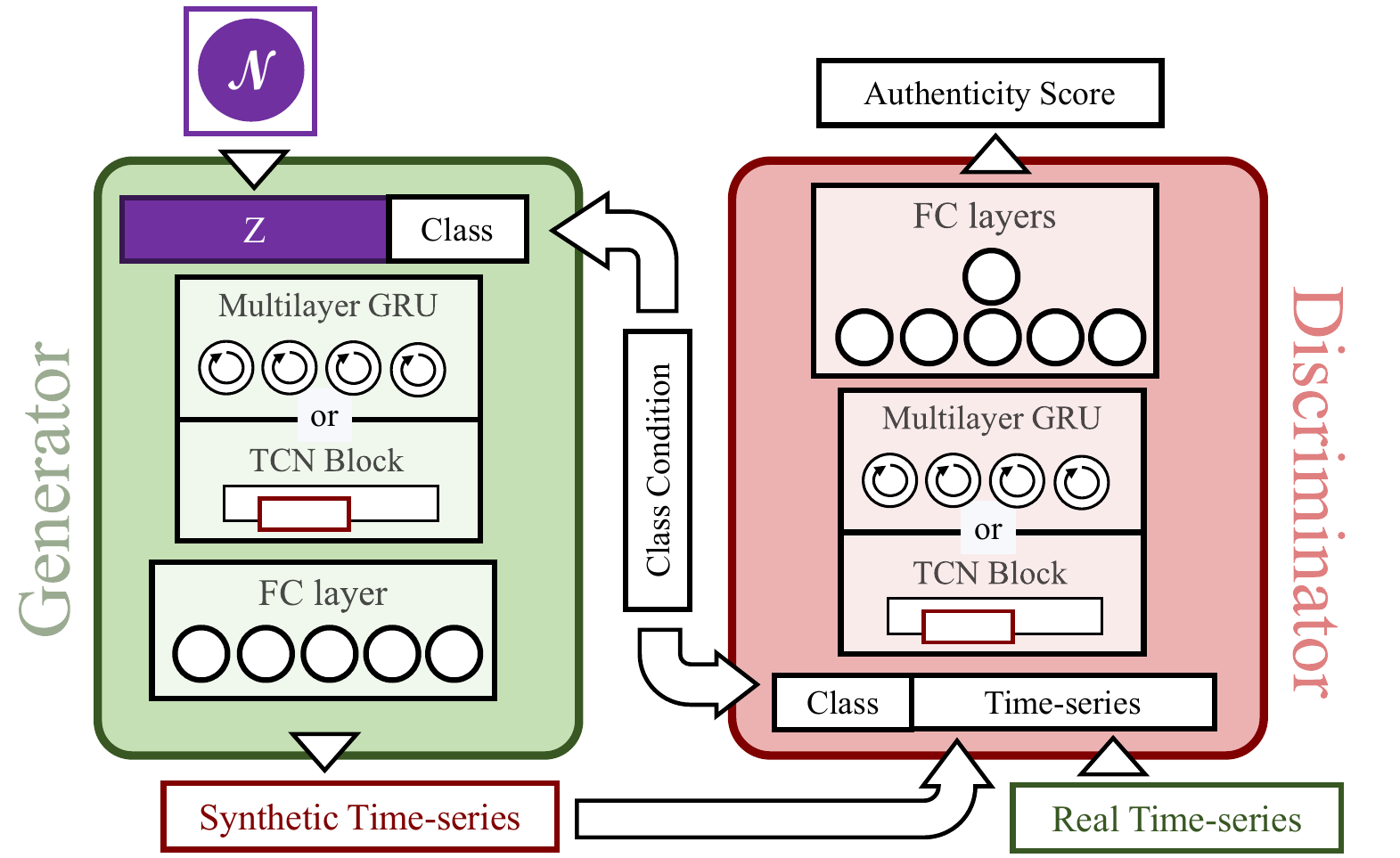}
    \caption{RCGAN architecture employed as the baseline in this study}
    \label{fig:rcgan_arch}
\end{figure}

To benchmark SNS-GAN's efficacy, we employ the Recurrent Conditional GAN (RCGAN)~\cite{esteban_real-valued_2017} as the baseline model. RCGAN leverages recurrent neural networks (RNNs) within the GAN architecture to generate time series data, primarily for medical applications. As depicted in Figure~\ref{fig:rcgan}, RCGAN feeds conditional information explicitly to both the generator and discriminator. The generator processes different noise vectors concatenated with the conditions at each time step (Figure~\ref{fig:rcgan_gen}), while the discriminator evaluates the authenticity of time series data at each individual step (Figure~\ref{fig:rcgan_dis}). 
\newline\newline\noindent
We adopt two RCGAN architectures, RCGAN-RNN, which adheres to the original specification using GRUs, and RCGAN-TCN, which substitutes RNN units with Temporal Convolutional Networks (TCNs) for its primary components. Figure~\ref{fig:rcgan_arch} presents the implemented RCGAN architecture, and Appendix~\ref{app:hp} details the RCGAN configurations.

\subsection{SNS-GAN for Class Conditional Time Series Generation}

\begin{figure}[htbp]
    \centering
    \includegraphics[width=\textwidth]{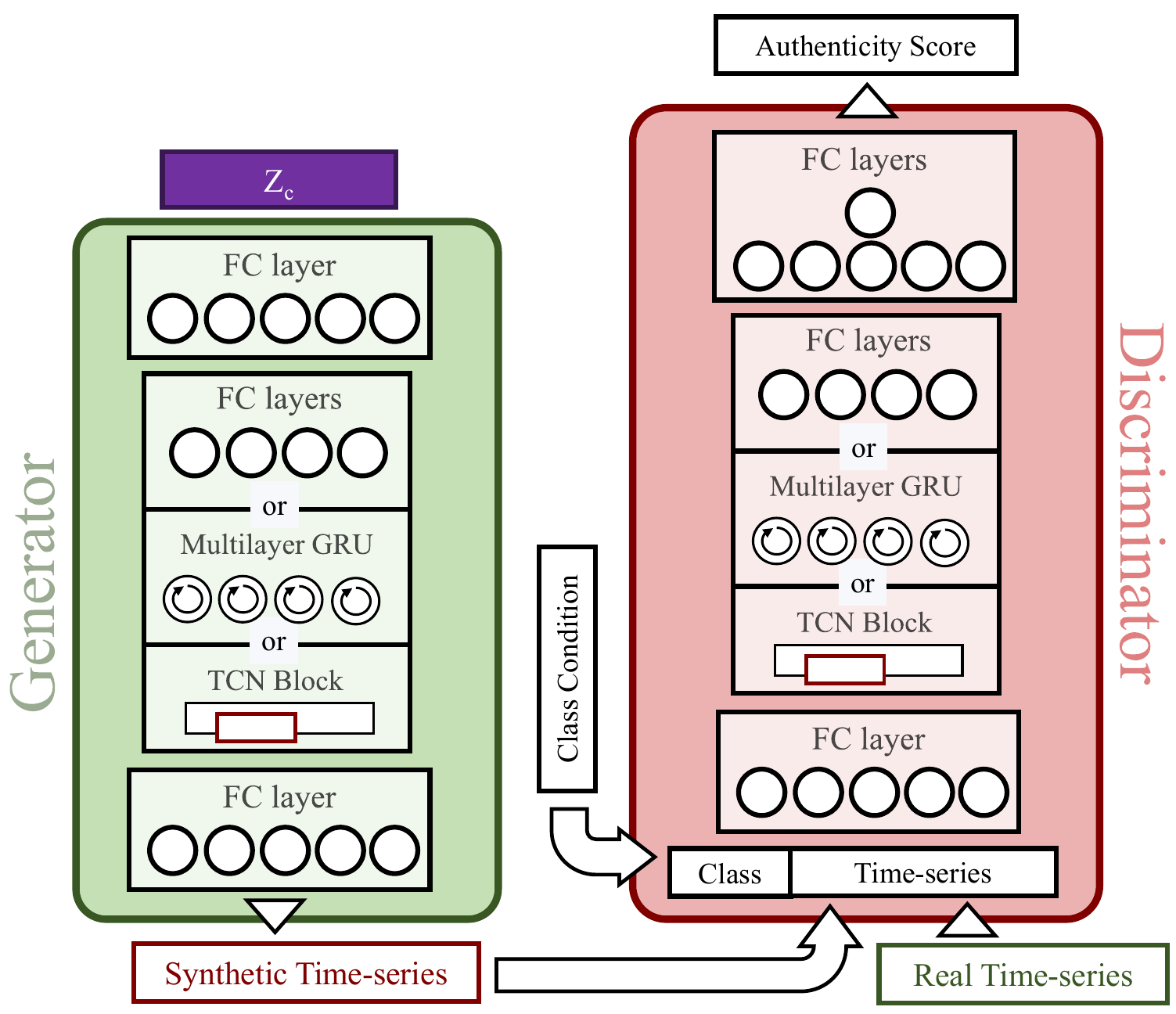}
    \caption{SNS-GAN architecture employed for time series generation}
    \label{fig:sns-gan_ts_arch}
\end{figure}

The SNS-GAN architecture tailored for time series is depicted in Figure~\ref{fig:sns-gan_ts_arch}. Three variants are introduced: SNS-GAN-Linear, which relies on fully connected layers; SNS-GAN-RNN, which uses GRUs; and SNS-GAN-TCN, which incorporates TCNs. The Linear and TCN variants generate entire sequences in one pass, whereas the RNN variant employs an autoregressive approach, where the time steps are generated sequentially. The hyperparameters for the SNS-GAN models are specified in Appendix~\ref{app:hp}.

\subsection{Results and Discussion}

\begin{figure}[h]
    \centering 
    \begin{subfigure}{\textwidth}
        \centering
        \includegraphics[width=0.33\textwidth]{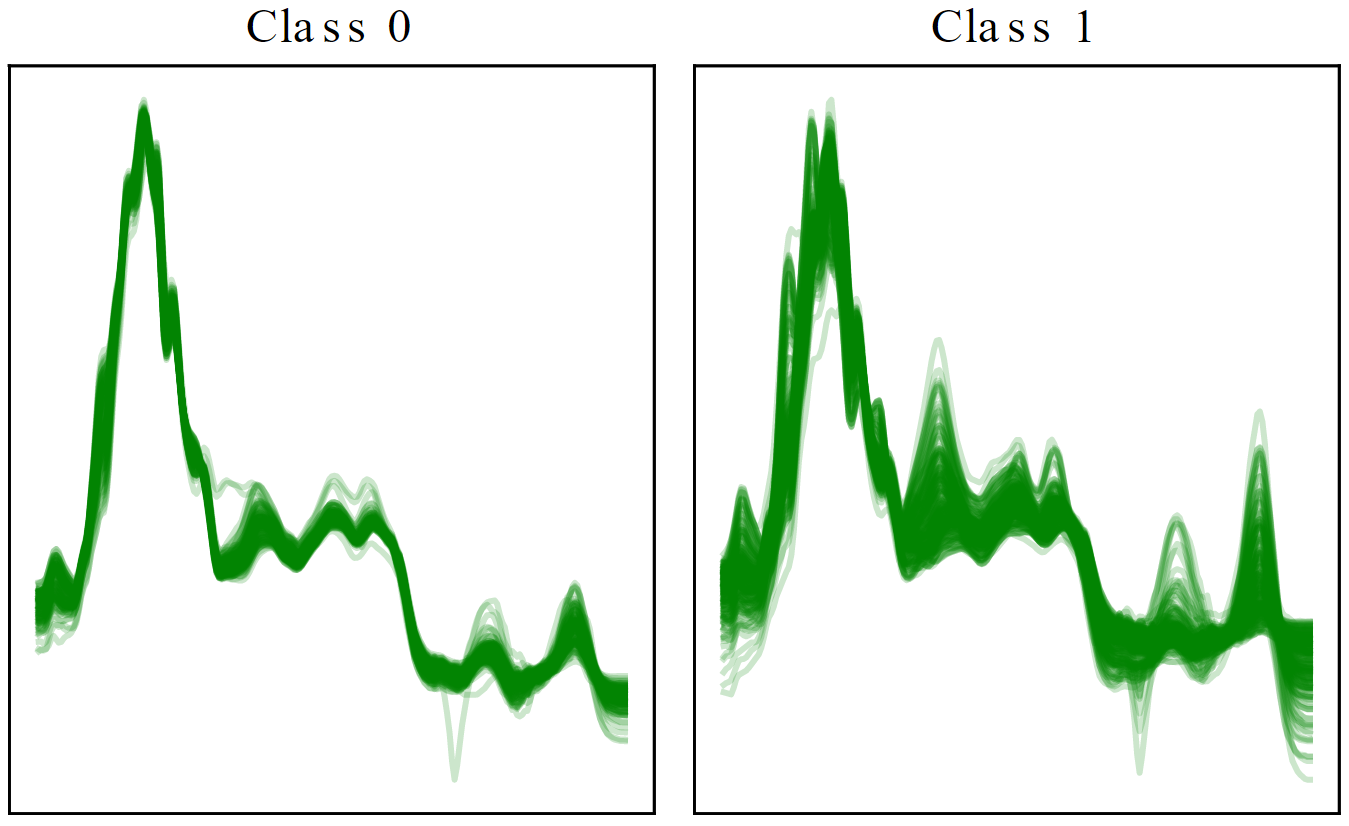}
        \caption{Real samples of Strawberry dataset}
    \end{subfigure}
    \vfill
    \begin{subfigure}{\textwidth}
        \centering
        \includegraphics[width=0.33\textwidth]{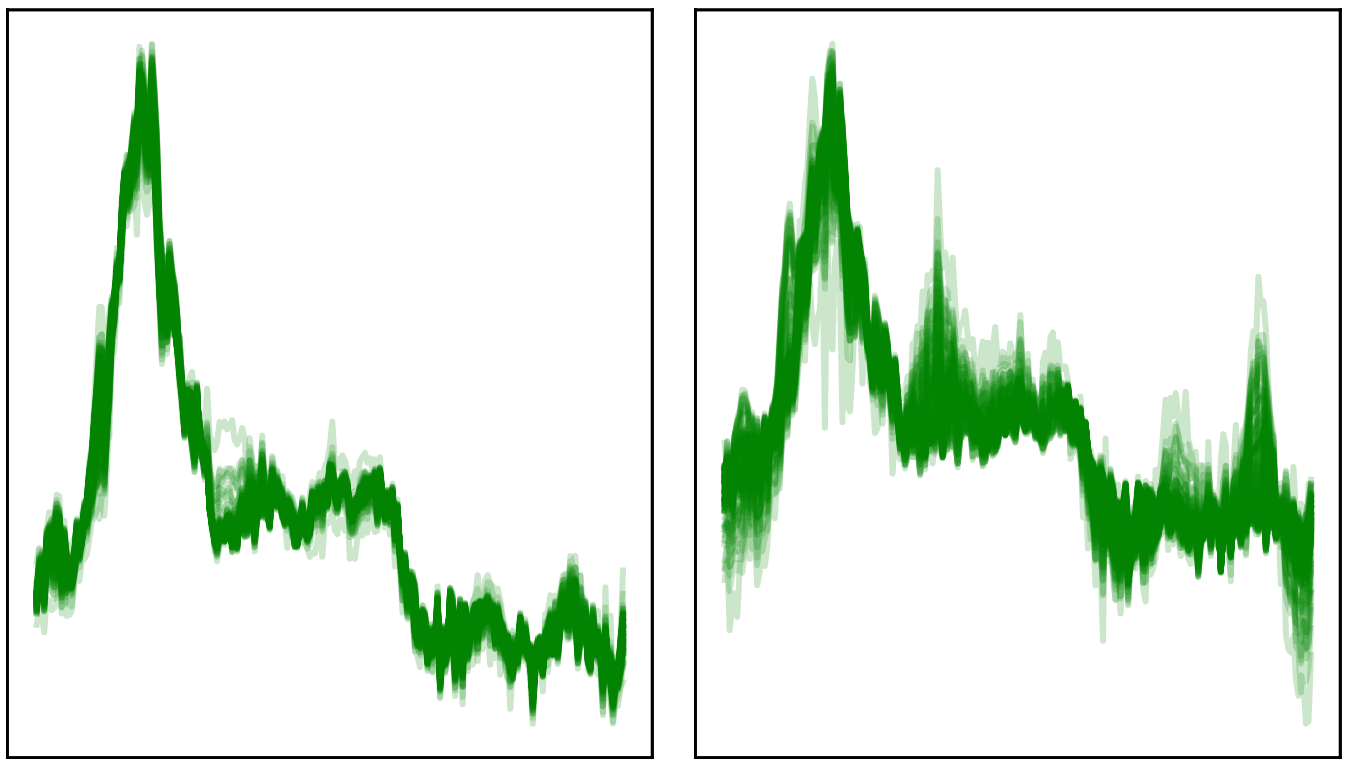}
        \caption{Generated samples for Strawberry dataset}
    \end{subfigure}
    \vfill
    \begin{subfigure}{0.55\textwidth}
        \includegraphics[width=\textwidth]{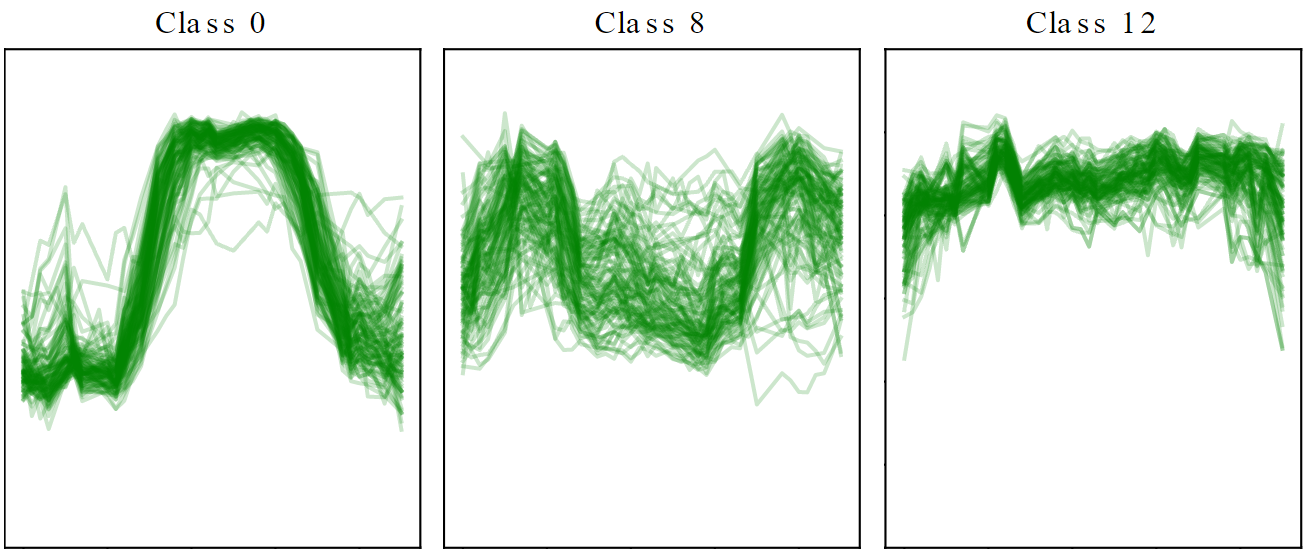}
        \caption{Real samples of Crop dataset}
    \end{subfigure}
    \vfill
    \begin{subfigure}{0.55\textwidth}
        \includegraphics[width=\textwidth]{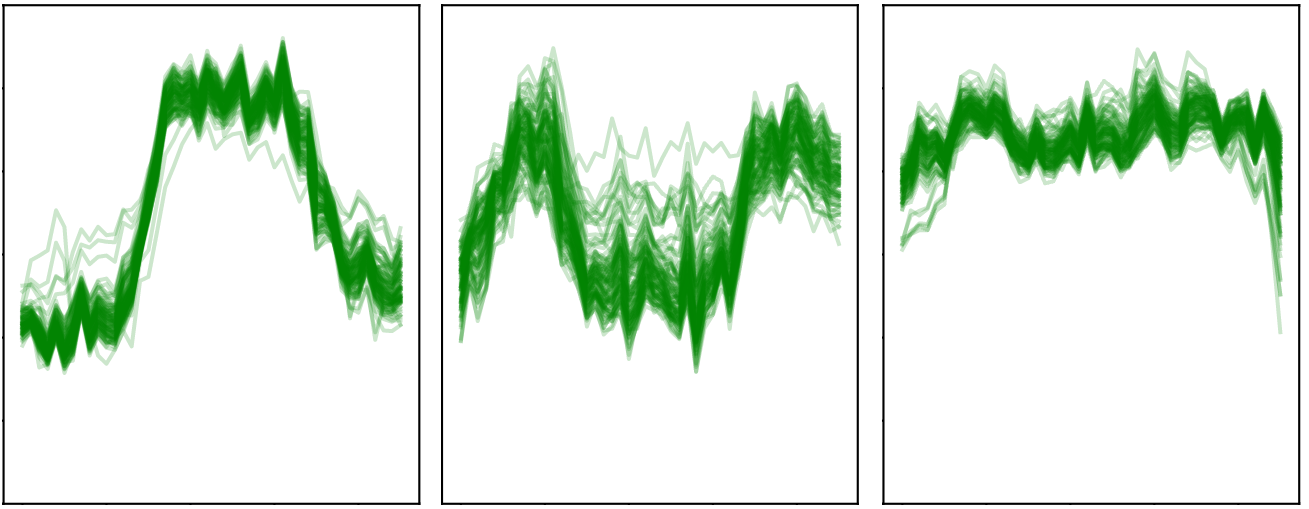}
        \caption{Generated samples of Crop dataset}
    \end{subfigure}
    \caption{Qualitative results for SNS-GAN-Linear model }
    \label{fig:sns_gan_linear_sample}
\end{figure}

To quantify the quality of generated samples in the time series domain, we utilized ITS and FITD~\cite{koochali_quantifying_2023}. The experimental outcomes are summarized in Table~\ref{tab:sns_gan_ts_res}. The SNS-GAN-Linear model stands out, demonstrating its capability to capture temporal dependencies across all time steps effectively. Figure~\ref{fig:sns_gan_linear_sample} showcases qualitative samples from this model, evidencing its accuracy in class representation and dynamic replication.
\begin{table}[h]
    \centering
        \caption{The quantitative results of class conditional time series generation}
    \label{tab:sns_gan_ts_res}
    \begin{tabular}{lcccc} \toprule
         &  Smooth Subspace & Strawberry & Crop & Fifty Words\\
         & ITS/FITD & ITS/FITD & ITS/FITD & ITS/FITD \\ \midrule
    Real Data & $\mathbf{2.93 / 0.097}$ & $\mathbf{1.98 / 0.039}$ & $\mathbf{22.76 / 0.0004}$ & $\mathbf{37.45 / 0.15}$ \\
    SNS-GAN-Linear & $\mathbf{2.88 / 2.04}$ & $\mathbf{1.967 / 0.30}$ & $\mathbf{19.19 / 0.05}$ & $\mathbf{35.08 / 1.65}$ \\
    SNS-GAN-TCN & $2.83 / 2.79$ & $1.961 / 0.82$ & $11.92 / 0.63$ & $10.77 / 4.48$ \\
    SNS-GAN-RNN & $2.66 / 3.73$ & $1.91 / 2.15$ & $18.88 / 0.07$ & $13.65 / 3.32$ \\
    RCGAN-TCN & $2.70 / 3.38$ & $1.95 / 1.17$ & $9.83 / 0.74$ & $8.49 / 23.49$ \\
    RCGAN-RNN & $2.59 / 4.45$ & $1.93 / 1.37$ & $15.86 / 0.14$ & $20.23 / 2.51$ \\ \bottomrule
    \end{tabular}
\end{table}
\newline\newline\noindent
While ITS scores are comparable for models on datasets with fewer classes, significant divergence is observed on datasets with a larger class spectrum. RNN-based models excel on the Crop dataset; however, their performance declines on the Fifty words dataset due to error accumulation in long series generation—contrasted by their proficiency with shorter series.
\newline\newline\noindent
The TCN-based model, however, does not distinctly excel on any dataset and is prone to mode collapse. Appendix~\ref{app:qual} presents visualizations of samples from the SNS-GAN models for further qualitative validation of results.
\newline\newline\newline\newline\noindent
In conclusion, the SNS-GAN-Linear variant emerges as the most adept at fulfilling condition imposition and generating authentic time series. It is best suited for fixed-size time series generation, while the RNN variant offers flexibility in length at the cost of diminished quality for longer sequences.

\section{Conclusion and Future Works}
In this paper, we introduced the Structured Noise Space GAN (SNS-GAN), a novel approach in class conditional generative modeling, particularly tailored for time series data. SNS-GAN stands out for its ability to incorporate class labels into the generative process without the need to alter the generator's architecture. This advancement marks a significant leap in the field, especially in handling class conditional generation in time series, a domain where traditional methods have shown limitations. Empirical evaluations demonstrate that SNS-GAN not only excels in image generation but shows remarkable effectiveness in the time series domain, surpassing baseline models by a considerable margin in various performance metrics.

Looking ahead, the potential of SNS-GAN extends beyond its current applications, offering fertile ground for future research. Its adaptability to different generative architectures without structural changes opens possibilities for its application across a wider array of datasets and in both image and time series domains. Moreover, the versatility of SNS-GAN indicates its potential applicability in other areas, such as music generation or synthetic voice creation. These future explorations promise to unveil the capabilities of SNS-GAN further, potentially revolutionizing the field of generative modeling across diverse data types. This not only underscores the innovative nature of SNS-GAN but also its transformative potential in advancing generative modeling techniques.

\clearpage
\bibliographystyle{unsrt}  
\bibliography{references}  

\appendixtitleon
\appendixtitletocon
\begin{appendices}
\section{Hyperparameters}
\label{app:hp}
This appendix provides the hyperparameters for various models adopted in this study. Tables~\ref{tab:app_sns_img_hp} and~\ref{tab:app_sns_gan_ts_hyp} present the hyperparameters of SNS-GAN models for image and time series data, respectively. Table~\ref{tab:app_rcgan_hyp} presents the hyperparameters of RCGAN in this study.

\begin{table}[h]
    \centering
    \caption{Hyperparameters for SNS-GAN in image domain experiments.}
    \label{tab:app_sns_img_hp}
    \begin{tabular}{@{}lcc@{}}
        \toprule
        & MNIST & CIFAR10 \\
        \midrule
        Noise size & 1000 (100$\times$10) & 1000 (100$\times$10) \\
        Transposed convolution layers & 2 & 4 \\
        Kernel size (Transposed convolution) & 4$\times$4 & 4$\times$4 \\
        Stride (Transposed convolution) & 2 & 2 \\
        Convolution layers (Discriminator) & 2 & 4 \\
        Kernel size (Convolution) & 4$\times$4 & 4$\times$4 \\
        Stride (Convolution) & 2 & 2 \\
        \bottomrule
    \end{tabular}
\end{table}

\begin{table}[h]
    \centering
    \caption{SNS-GAN hyperparameters}
    \label{tab:app_sns_gan_ts_hyp}
    \begin{tabular}{llll} \toprule
        & Variant & Generator & Discriminator \\ \midrule
    FC layers & SNS-GAN-Linear& 1 & 1 \\
    FC hidden size & SNS-GAN-Linear& 500 & 500 \\
    GRU layers & SNS-GAN-RNN& 1 & 1 \\
    GRU hidden size & SNS-GAN-RNN& 512 & 512 \\
    TCN layers & SNS-GAN-TCN& 1 & 1 \\
    TCN kernel size & SNS-GAN-TCN& 8 & 8 \\ \bottomrule
    \end{tabular}
\end{table}

\begin{table}[h]
    \centering
    \caption{RCGAN hyperparameters}
    \label{tab:app_rcgan_hyp}
    \begin{tabular}{llll} \toprule
        & Variant & Generator & Discriminator \\ \midrule
    GRU layers & RCGAN-RNN& 1 & 1 \\
    GRU hidden size & RCGAN-RNN& 256 & 256 \\
    TCN layers & RCGAN-TCN& 1 & 1 \\
    TCN kernel size & RCGAN-TCN& 8 & 8 \\ \bottomrule
    \end{tabular}
\end{table}

\clearpage

\section{Further Qualitative Results}
\label{app:qual}
\begin{figure}[h]
    \centering 
    \begin{subfigure}{0.55\textwidth}
        \includegraphics[width=\textwidth]{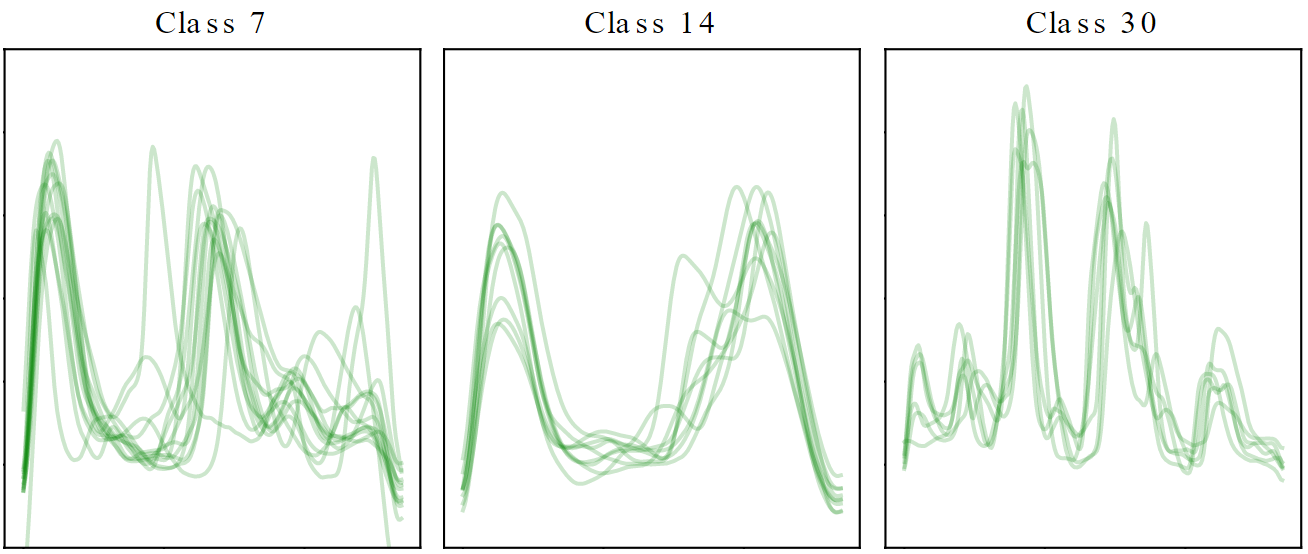}
        \caption{Real samples of Fifty words dataset}
    \end{subfigure}
    \vfill
    \begin{subfigure}{0.55\textwidth}
        \includegraphics[width=\textwidth]{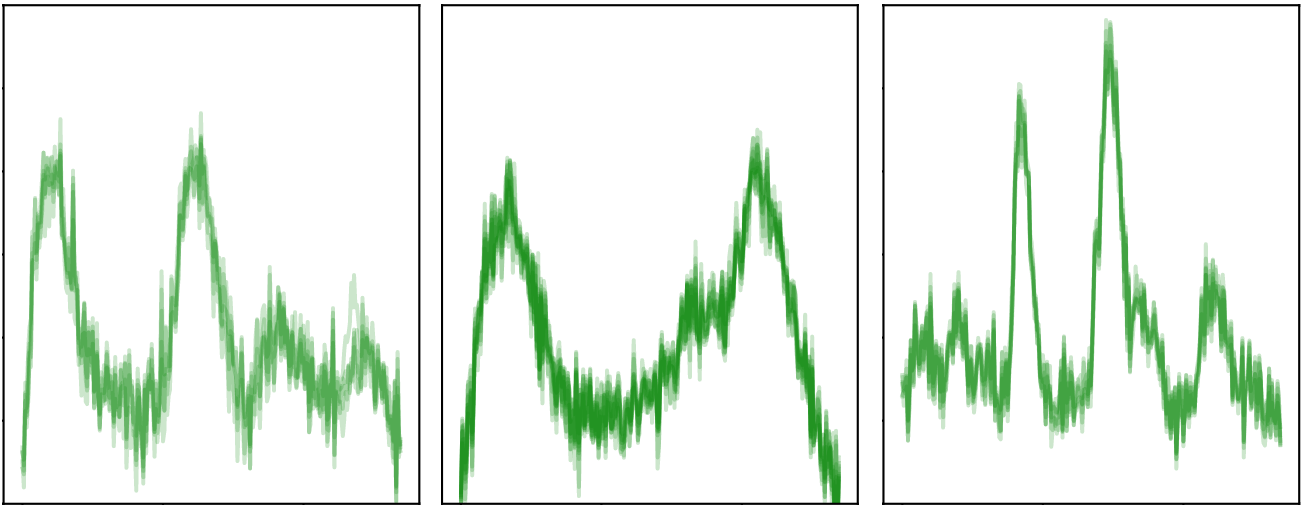}
        \caption{Generated samples of Fifty words dataset}
    \end{subfigure}
    \caption{Qualitative results for SNS-GAN-Linear model }
    \label{fig:app_sns_gan_linear_sample}
\end{figure}

\begin{figure}
    \centering 
    \begin{subfigure}{\textwidth}
        \centering
        \includegraphics[width=0.33\textwidth]{sns_gan/real_strawberry.png}
        \caption{Real samples of Strawberry dataset}
    \end{subfigure}
    \vfill
    \begin{subfigure}{\textwidth}
        \centering
        \includegraphics[width=0.33\textwidth]{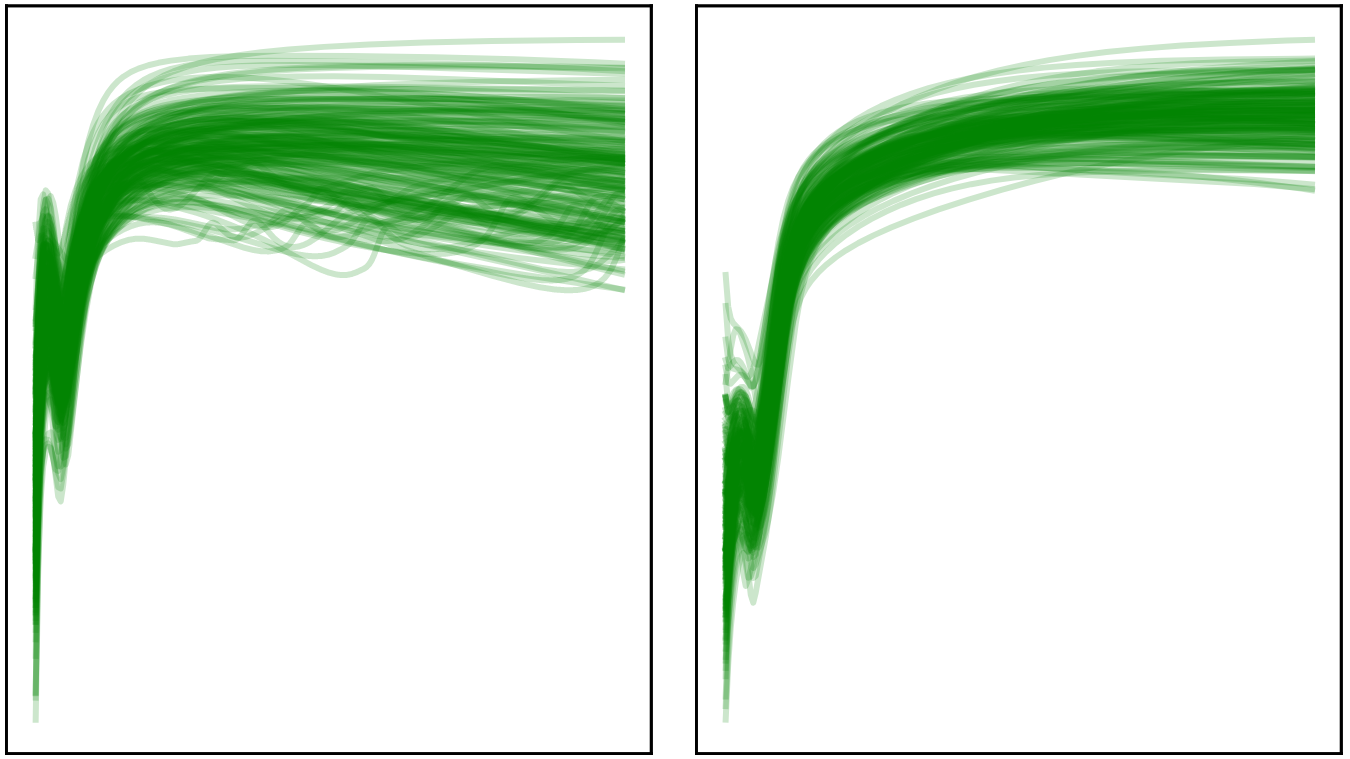}
        \caption{Generated samples for Strawberry dataset}
    \end{subfigure}
    \vfill
    \begin{subfigure}{0.55\textwidth}
        \includegraphics[width=\textwidth]{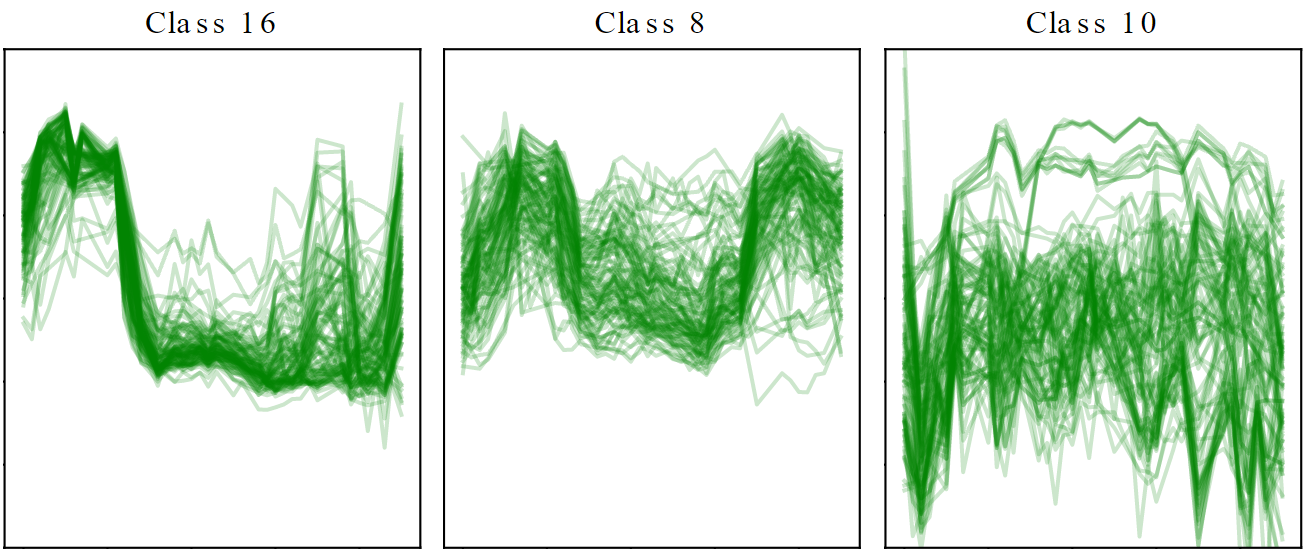}
        \caption{Real samples of Crop dataset}
    \end{subfigure}
    \vfill
    \begin{subfigure}{0.55\textwidth}
        \includegraphics[width=\textwidth]{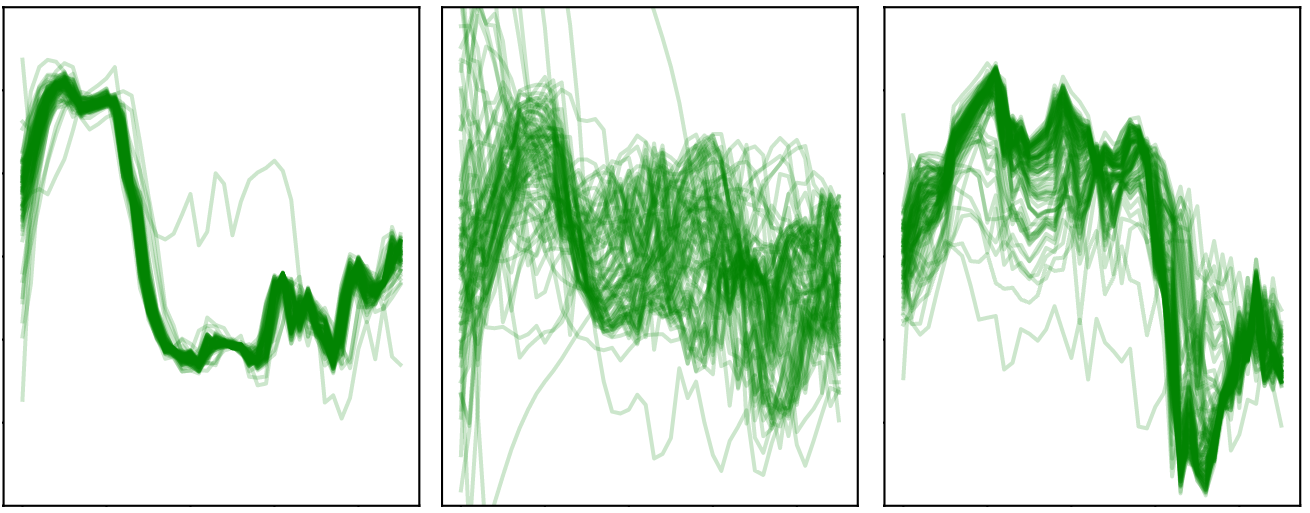}
        \caption{Generated samples of Crop dataset}
    \end{subfigure}
    \caption{Qualitative results for SNS-GAN-RNN model }
    \label{fig:app_sns_gan_rnn_sample}
\end{figure}

\begin{figure}
    \centering 
    \begin{subfigure}{\textwidth}
        \centering
        \includegraphics[width=0.33\textwidth]{sns_gan/real_strawberry.png}
        \caption{Real samples of Strawberry dataset}
    \end{subfigure}
    \vfill
    \begin{subfigure}{\textwidth}
        \centering
        \includegraphics[width=0.33\textwidth]{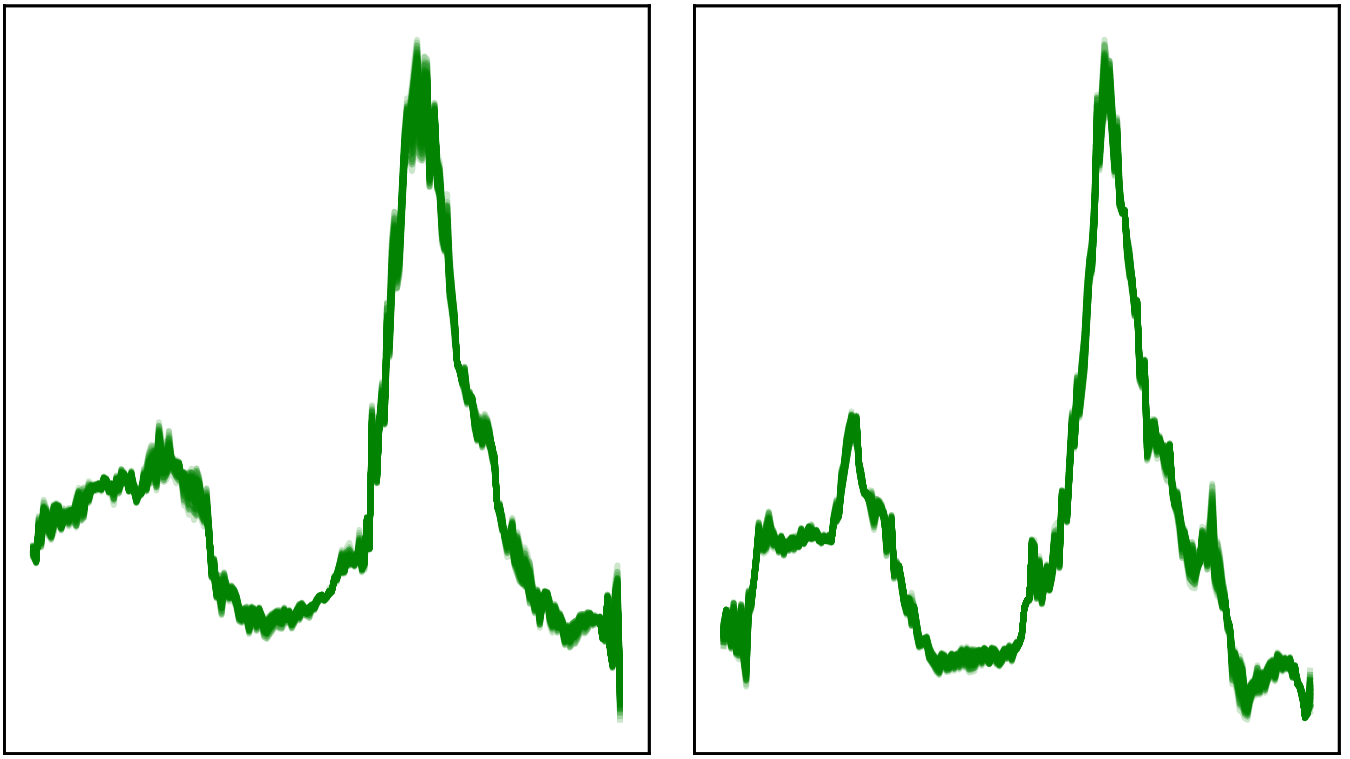}
        \caption{Generated samples for Strawberry dataset}
    \end{subfigure}
    \vfill
    \begin{subfigure}{0.55\textwidth}
        \includegraphics[width=\textwidth]{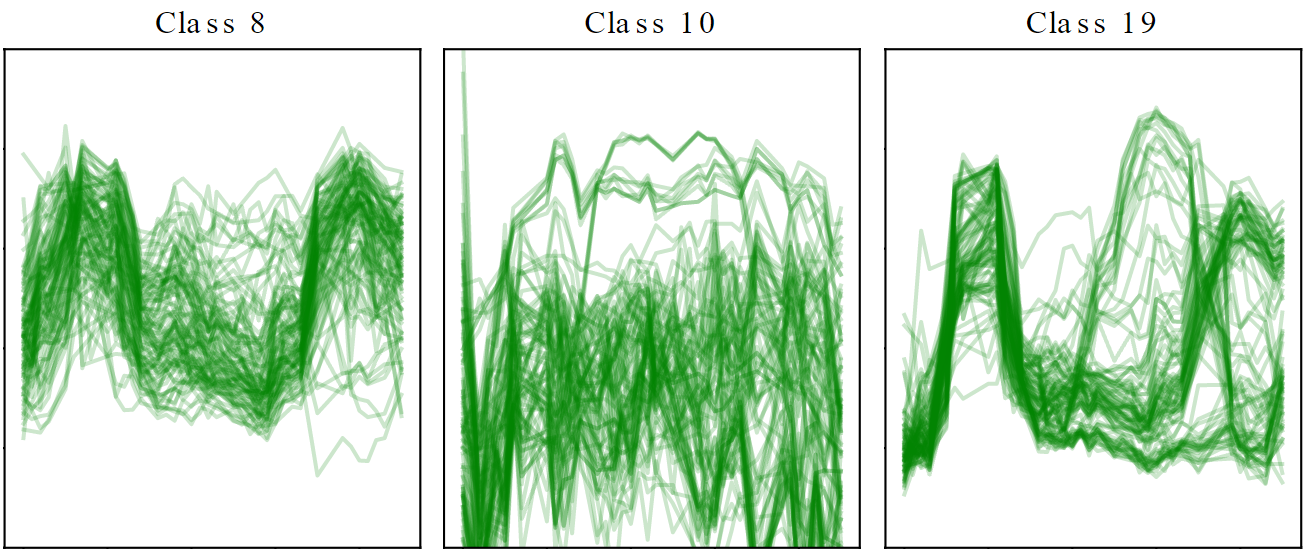}
        \caption{Real samples of Crop dataset}
    \end{subfigure}
    \vfill
    \begin{subfigure}{0.55\textwidth}
        \includegraphics[width=\textwidth]{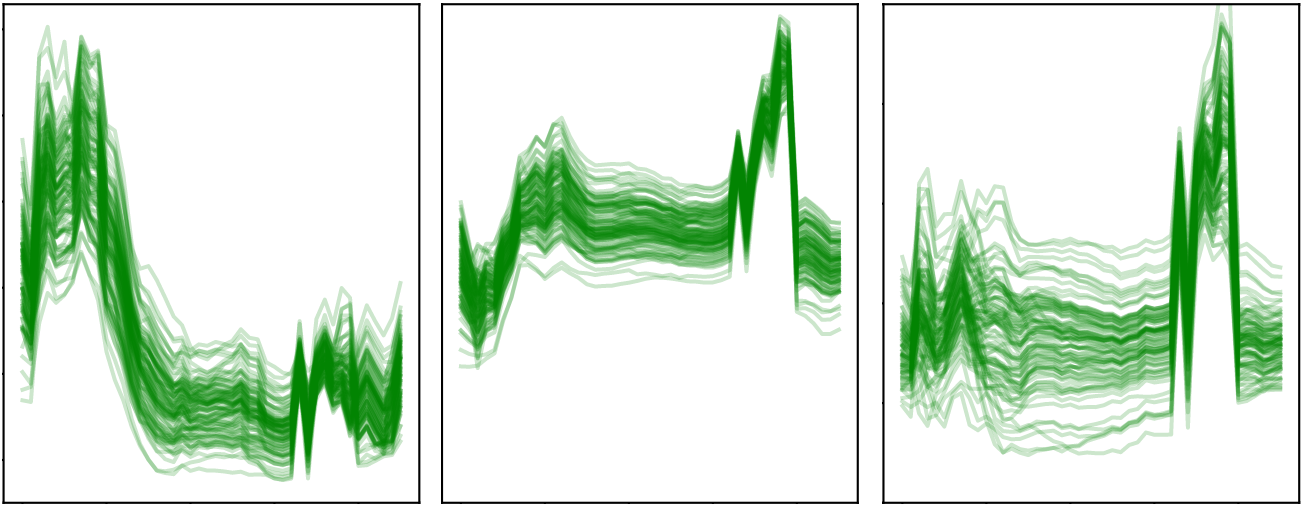}
        \caption{Generated samples of Crop dataset}
    \end{subfigure}
    \caption{Qualitative results for SNS-GAN-TCN model }
    \label{fig:app_sns_gan_tcn_sample}
\end{figure}

This appendix provides further visualization of samples generated by various SNS-GAN architectures. Figure~\ref{fig:app_sns_gan_linear_sample} depicts samples SNS-GAN-Linear. Figure~\ref{fig:app_sns_gan_rnn_sample} illustrates samples from SNS-GAN-RNN and figure~\ref{fig:app_sns_gan_tcn_sample} presents samples from SNS-GAN-TCN.

\end{appendices}
\end{document}